\def\year{2022}\relax
\documentclass[letterpaper]{article} 
\usepackage{aaai22}  
\usepackage{times}  
\usepackage{helvet} 
\usepackage{courier}  
\usepackage[hyphens]{url}  
\usepackage{graphicx} 
\usepackage{natbib}  
\usepackage{caption} 
\usepackage{booktabs}
\usepackage{amsfonts} 
\usepackage{amsmath}
\usepackage{bm}
\usepackage{tabularx}

\title{CEM: Commonsense-aware Empathetic Response Generation}
\author {
    Sahand Sabour, Chujie Zheng, Minlie Huang \\
}
\affiliations{
  The CoAI Group, DCST, Institute for Artificial Intelligence, \\
  State Key Lab of Intelligent Technology and Systems, \\
  Beijing National Research Center for Information Science and Technology, \\
  Tsinghua University, Beijing 100084, China \\
  {\tt \{sahandfer, chujiezhengchn\}@gmail.com, aihuang@tsinghua.edu.cn} \\
}
\begin{document}

\maketitle

\begin{abstract}
A key trait of daily conversations between individuals is the ability to express empathy towards others, and exploring ways to implement empathy is a crucial step towards human-like dialogue systems. 
Previous approaches on this topic mainly focus on detecting and utilizing the user's emotion for generating empathetic responses.
However, since empathy includes both aspects of affection and cognition, we argue that in addition to identifying the user's emotion, cognitive understanding of the user's situation should also be considered. 
To this end, we propose a novel approach for empathetic response generation, which leverages commonsense to draw more information about the user's situation and uses this additional information to further enhance the empathy expression in generated responses.
We evaluate our approach on \textsc{EmpatheticDialogues}, which is a widely-used benchmark dataset for empathetic response generation. 
Empirical results demonstrate that our approach outperforms the baseline models in both automatic and human evaluations and can generate more informative and empathetic responses.
Our code is available from \url{https://github.com/Sahandfer/CEM}. 
\end{abstract}

\section{Introduction}
Empathy is a desirable trait of human daily conversations that enables individuals to understand, perceive, and respond appropriately to the situation and feelings of others \cite{Keskin2014}. Previous research has demonstrated that empathetic dialogue systems can improve user experience and satisfaction in multiple domains \cite{fitzpatrick2017emp, liu2021emotional, wang2021cass}. Hence, it is important to discover ways that allow us to equip open-domain dialogue systems with empathy. Recent work \cite{rashkin2019empathetic, lin2019moel, majumder2020mime, li2020empdg} has proposed various methods of generating empathetic responses that mainly rely on detecting the user's emotion. 

\begin{figure}[ht]
    \centering
    \includegraphics[width=\linewidth]{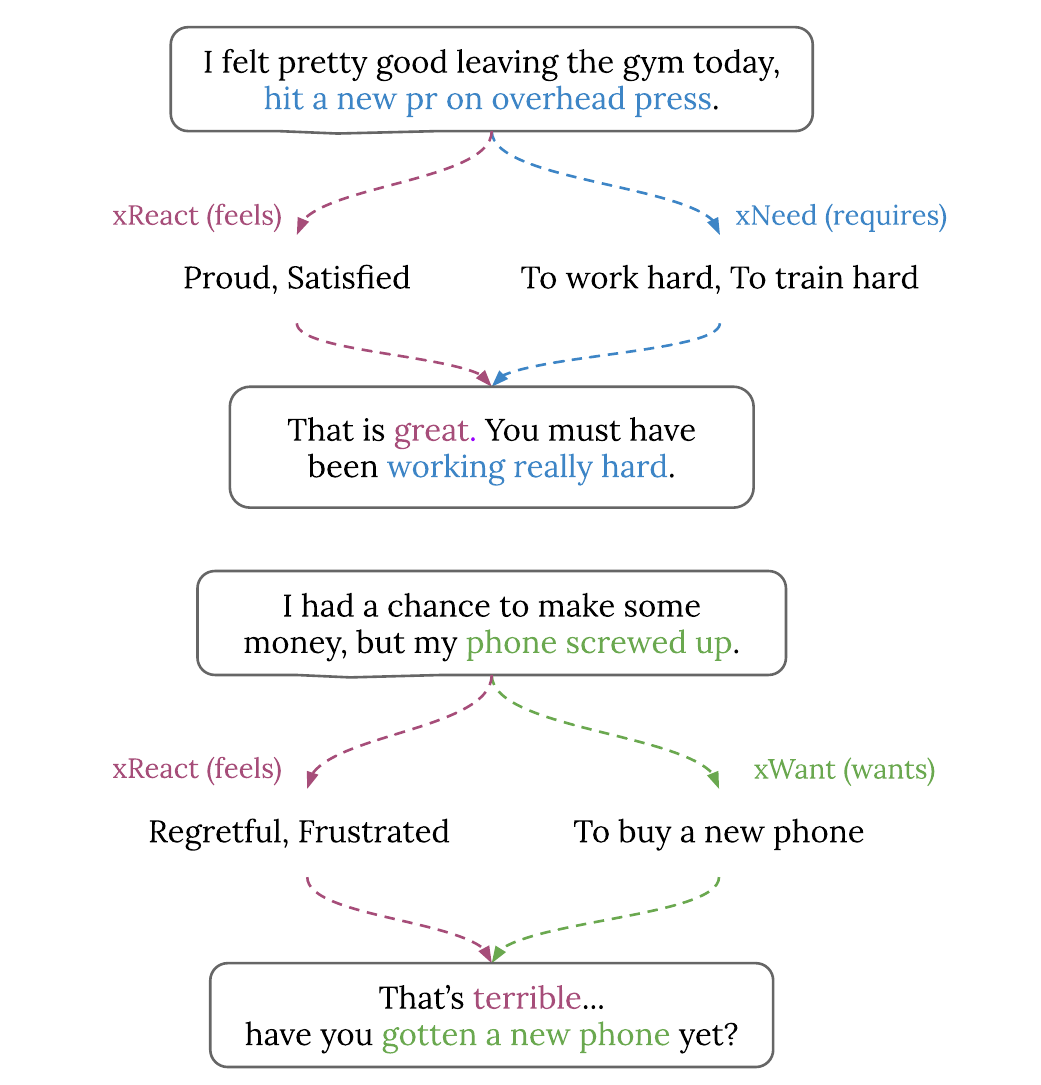}
    \caption{Examples from the \textsc{EmpatheticDialogues} dataset in which commonsense is used to gain additional information about the user's emotion and situation before responding empathetically.}
    \label{fig:intro}
\end{figure}

However, empathy is a broad construct that includes aspects of affection and cognition \cite{davis1983empathy}. The affective aspect is concerned with the emotional simulation in reaction to the user's experiences \cite{cuff2016empathy} while the cognitive aspect aims to understand the user's situation and the implied feelings \cite{elliott2018empathy}. Hence, though emotion is one of the important factors of empathy, it is not the only determining factor. This is demonstrated in Figure \ref{fig:intro}, where both affective and cognitive empathy are used to form empathetic responses. For instance, in the first case, the user shares information about their emotion (\textit{I felt pretty good}) as well as their experience (\textit{hit a new pr on the overhead press}). Accordingly, we can assume that the user is \textit{Proud} of their achievement and must have \textit{worked hard} to reach this level. Since these assumptions are not explicitly mentioned by the user, we as humans tend to rely on our own knowledge and commonsense reasoning to draw these implications. Therefore, we believe that providing dialogue systems with this external knowledge could play a critical role in understanding the user's situation and feelings, which leads to more informative and empathetic responses. 

Towards this end, we propose the \textbf{C}ommonsense-aware \textbf{E}mpathetic Chatting \textbf{M}achine (\textbf{CEM}).
CEM leverages external commonsense knowledge to obtain more information about the user's situation and feelings (i.e. user's reaction, intention, desire, etc.).
Such additional information is used to improve the cognitive understanding and thus, enhance the empathy expression in the generated responses. 
We evaluate our approach on \textsc{EmpatheticDialogues}, a widely-used benchmark dataset for empathetic response generation. 
Both automatic and manual evaluation results demonstrate that compared to previous methods, CEM can generate more informative and empathetic responses.

Our contributions are summarized as follows:
\begin{itemize}
    \item We propose to leverage commonsense to improve the understanding of interlocutors' situations and feelings, which is an important part of cognitive empathy.
    \item We introduce CEM, a novel approach that uses various types of commonsense reasoning to enhance empathetic response generation.
    \item Automatic and manual evaluation demonstrate that with the addition of commonsense, CEM is able to generate more informative and empathetic responses compared with the previous methods.
\end{itemize}

\section{Preliminaries}
\subsection{Empathetic Dialogue Generation}
Empathy is a fairly new term in the literature and therefore, has no specific or widely accepted definition in the fields of social psychology and psychotherapy \cite{macarov1978empathy, elliot2011empathy}. However, empathy is commonly known as a complex multi-dimensional construct that includes broad aspects of affection and cognition \cite{davis1983empathy, zheng-etal-2021-comae}. 
Affective empathy enables us to experience the emotion of others through various emotional stimuli \cite{cuff2016empathy}, while cognitive empathy enables us to understand the situations and implicit mental states of others, such as intentions, causes, desires, requirements, etc. \cite{elliott2018empathy}.

In recent years, research on implementing empathy in dialogue systems and generating empathetic responses has gained considerable attention. Initially, \citet{rashkin2019empathetic} demonstrated that detecting the user's emotion is an essential part of generating empathetic responses. \citet{lin2019moel} designed a separate decoder for each available emotion and softly combined their outputs. \citet{majumder2020mime} proposed that empathetic responses should also mimic the user's emotion to a degree. \citet{li2020empdg} leveraged user feedback and proposed a multi-resolution adversarial framework for this task. Recently, \cite{li2020empathetic} used commonsense knowledge from ConceptNet \cite{speer2018conceptnet} to gain a better understanding of the implied emotions within the context. However, these works usually focus on detecting the context emotion and do not pay enough attention to the cognitive aspect of empathy.

\subsection{Commonsense and Empathy}
As mentioned, a major part of cognitive empathy is understanding the situations and feelings of others. 
When interacting with a dialogue system, the user is not expected to explicitly share all the information about their situation and how they may feel. As humans, we use our commonsense knowledge to make connections between what is explicitly mentioned and what is implied. Hence, we hypothesize that enabling dialogue systems to leverage commonsense and drive implications from what the user has explicitly shared is highly beneficial for a better understanding of the user's situation and feelings, which leads to more effective cognitive empathy and thus, more empathetic responses.

In this work, we use ATOMIC \cite{sap2019atomic} as our commonsense knowledge base. 
ATOMIC is a collection of commonsense reasoning inferences about everyday if-then events. 
For each event, ATOMIC infers six commonsense relations for the person involved in the event:
the effect of the event on the person (\textit{xEffect}), their reaction to the event (\textit{xReact}), their intent before the event (\textit{xIntent}), what they need in order for the event to happen (\textit{xNeed}), what they would want after the event(\textit{xWant}), and an inferred attribute of the person's characteristics (\textit{xAttr}). Since predicting a person's attributes merely based on a given event would include judging the other person, which is not included in the empathetic process \cite{peloquin1995empathy}, we neglect \textit{xAttr} in our approach and use the remaining five relations. 

In order to generate commonsense inferences for given events, we adopt COMET \cite{bosselut2019comet}, which is a pre-trained GPT-2 model \cite{Radford2018ImprovingLU} that is finetuned on triplets $(e, r, i)$ from ATOMIC, where $e, r, i$ are the event, the relation type, and the inferred knowledge respectively. More specifically, we use a modified BART-based \cite{lewis2019bart} variation of COMET, which is trained on the ATOMIC-2020 dataset \cite{hwang2020cometatomic}. This model is equipped with knowledge that is not readily available to pre-trained language models and is more suitable for inferring knowledge regarding unseen events \cite{hwang2020cometatomic}. The latter is necessary for our use-case as many of the events within an empathetic conversation may not occur on a daily basis and therefore, may not exist in the original ATOMIC dataset.

\begin{figure*}[t]
    \centering
    \includegraphics[width=\linewidth]{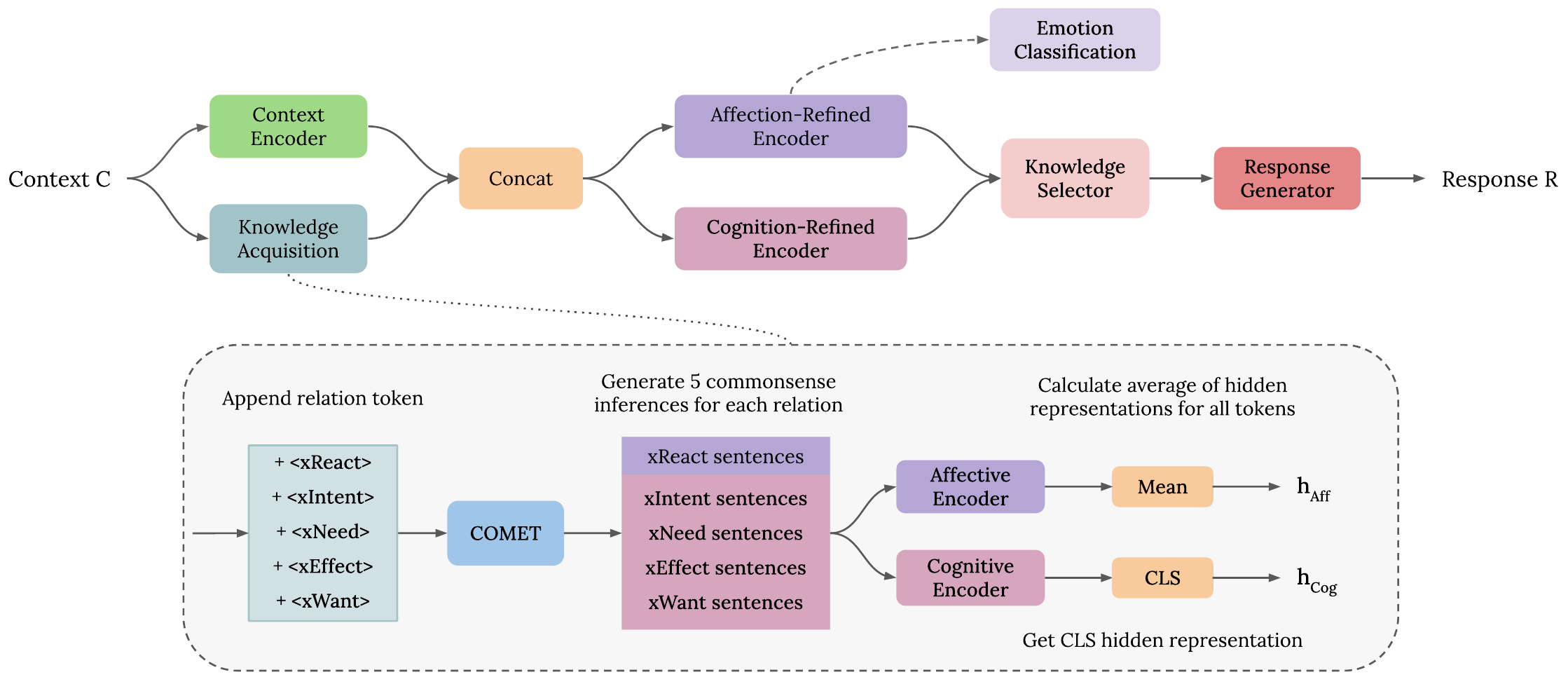}
    \caption{Overview of our model (CEM).}
    \label{fig:CEM}
\end{figure*}

\subsection{Task Formulation}
\label{subsec:formulation}
We conduct our experiments on the \textsc{EmpatheticDialogues} \cite{rashkin2019empathetic}, a large-scale multi-turn dataset containing 25k empathetic conversations between crowdsourcing workers. The dataset also provides an emotion label for each conversation from the total 32 available emotions.
In this dataset, each conversation is between a speaker and a listener.
The task requires a dialogue model to play the role of the listener and generate empathetic responses.
Formally, let $D = [u_1, u_2, u_3, ..., u_{k-1}]$ denote a dialogue history of $k-1$ utterances, where $u_i = [w^i_1, w^i_2, w^i_3, ..., w^i_{M_i}]$ is the $i$-th utterance that consists of $M_i$ words.
Our goal is to generate the listener's next utterance $u_k$ which is coherent to the context, informative, and empathetic to the speaker's situation and feelings.

\section{Methodology}
Our proposed model, CEM, is built upon the standard Transformer \cite{vaswani2017attention} and its overview is illustrated in Figure \ref{fig:CEM}. The process of CEM is mainly divided into five stages: context encoding, knowledge acquisition,  context refinement, knowledge selection, and response generation.

\subsection{Context Encoding}
Following previous work \cite{lin2019moel, majumder2020mime}, we concatenate the utterances in the dialogue history and prepend a special token \texttt{[CLS]} to obtain the context input $C=\texttt{[CLS]} \oplus u_1\oplus u_2\oplus u_3\oplus ... \oplus u_{k-1}$, where $\oplus$ is the concatenation operation. 
Similar to \citet{devlin2019bert}, we use the final hidden representation of \texttt{[CLS]} as the representation of the whole sequence.

We acquire the embedding $\bm{E}_C$ of the sequence $C$ by summing up the word embedding, positional embedding, and dialogue state embedding. As each utterance in $C$ could be from either the listener or the speaker, we use the dialogue state embedding to distinguish between the two parties.
The sequence embedding $\bm{E}_C$ is then fed to a context encoder to produce the contextual representation:
\begin{equation}
    \label{eqn:init}
    \bm{H}_{CTX} = \mathbf{Enc}_{CTX}(\bm{E}_C)
\end{equation}
where $\bm{H}_{CTX} \in \mathbb{R}^{L\times d}$, $L$ is the length of the sequence, and $d$ is the hidden size of the context encoder. 

\subsection{Knowledge Acquisition}
For input sequence $C$, we respectively append five special relation tokens (\texttt{[xReact]}, \texttt{[xWant]}, \texttt{[xNeed]}, \texttt{[xIntent]}, \texttt{[xEffect]}) to the last utterance in the dialogue history and use COMET to generate five commonsense inferences $[cs^{r}_1,cs^{r}_2, ..., cs^{r}_5]$ per relation $r$. 
For each relation, we concatenate the generated commonsense inferences to obtain its commonsense sequence $CS_{r} = cs^{r}_1\oplus cs^{r}_2\oplus ...\oplus cs^{r}_5$. 
Given that xReact demonstrates the knowledge regarding the affective state (i.e. user's emotion) while the other relation represents knowledge regarding the cognitive state (i.e. user's situation), we divide the relations into two groups: affective and cognitive. 
Accordingly, similar to the previous section, we prepend \texttt{[CLS]} to the cognitive sequences. As the inferences for xReact are usually emotion words (e.g. \textit{sad, happy, angry}) rather than sentences, we would simply use the average of its hidden representations to represent these sequences. Based on the mentioned grouping, the resulting sequences are fed to two separate cognitive and affective encoders:
\begin{align}
     \bm{H}_\textit{xReact} &= \mathbf{Enc}_{Aff}(\bm{E}_{CS_\textit{xReact}}) 
    \label{eqn:aff}\\
     \bm{H}_{r} &= \mathbf{Enc}_{Cog}(\bm{E}_{CS_{r}}) 
    \label{eqn:cog}
\end{align}
where $\bm{H}_\textit{xReact} \in \mathbb{R}^{l_\textit{xReact}\times d}, \bm{H}_{r} \in \mathbb{R}^{l_r\times d}$, with $l_\textit{xReact}, l_r$ being the lengths of the commonsense inference sequences, and $ r \in $\{\textit{xWant}, \textit{xNeed},  \textit{xIntent},  \textit{xEffect}\}. 

Then, we use the average hidden representation for affective relations and the hidden representation of \texttt{[CLS]} for cognitive relations to represent these sequences respectively:
\begin{align}
     \bm{h}_\textit{xReact} &= \mathrm{Average}(\bm{H}_\textit{xReact})
     \label{eqn:1} \\
     \bm{h}_{r} &= \bm{H}_{r}[0]
     \label{eqn:2}
\end{align}
where $h_\textit{xReact}, h_{r} \in \mathbb{R}^{d}$. 

\subsection{Context Refinement}
Similar to \citet{majumder2020mime}, in order to refine the context by additional information, we first respectively concatenate each of the commonsense relation representations (Equations \ref{eqn:1} \& \ref{eqn:2}) to the context representation $\bm{H}_{CTX}$ at the token level (i.e. $\bm{U}_{r} \in \mathbb{R}^{L\times2d}$):
\begin{align}
    \bm{U}_\textit{xReact}[i] &= \bm{H}_{CTX}[i] \oplus \bm{h}_\textit{xReact} \\
    \bm{U}_{r}[i] &= \bm{H}_{CTX}[i] \oplus \bm{h}_{r}
\end{align}
In contrast to concatenating the representations at a sequence level (i.e. adding additional information to the end of the context representation), token-level concatenation enables us to fuse the additional knowledge within each word in the sequence. 

Accordingly, we use two separate encoders (affection-refined and cognition-refined), corresponding to the two groups of relations, to encode the fused representations and obtain commonsense-refined context representations for each relation respectively:
\begin{equation}
    \label{eqn:7}
    \bm{H}_{Aff} = \mathbf{Enc}_{CTX-Aff}(\bm{U}_\textit{xReact})
\end{equation}
\begin{equation}
    \label{eqn:8}
    \bm{H}_{Cog, r} = \mathbf{Enc}_{CTX-Cog}(\bm{U}_{r})
\end{equation}
where $\bm{H}_{Aff}, \bm{H}_{Cog, r} \in \mathbb{R}^{L\times d}$.

\subsubsection{Emotion Classification} In order to acquire a more accurate prediction of the user's affective state, given that we are provided with an emotion label $e^*$ for each conversation, we use the hidden representation of the \texttt{[CLS]} token from the affection-refined context representation ($\bm{h}_{Aff}$) to perform emotion classification: 
\begin{equation}
    \bm{h}_{Aff} = \bm{H}_{Aff}[0]
\end{equation}
where $\bm{h}_{Aff} \in \mathbb{R}^{d}$. Hence, we pass $\bm{h}_{Aff} $ through a linear layer followed by a Softmax operation to produce the emotion category distribution $P_{emo} \in \mathbb{R}^{q}$, where $q$ is the number of available emotion categories:
\begin{equation}
    P_{emo} = \mathrm{Softmax}(\bm{W}_e \bm{h}_{Aff})
\end{equation}
where $\bm{W}_e \in \mathbb{R}^{d\times q}$ is the weight vector for the linear layer. During training, we optimize these weights by minimizing the Cross-Entropy (CE) loss between the emotion category distribution $P$ and the ground truth label $e^*$:
\begin{equation}
    \mathcal{L}_{emo} = - \log(P_{emo}(e^*))
\end{equation}

\subsection{Knowledge Selection}
Merely using one of the commonsense representations to produce an empathetic response is not ideal. For instance, if we only rely on the affection-refined context, the generated responses would likely be about how the user's emotions (e.g. \textit{You must be proud.}), whereas using the cognition-refined contexts may lead to responses that focus more on the situation (e.g. \textit{You must have worked really hard.}). Hence, we want to enable our model to generate responses based on the mixture of both affective and cognitive information. To this end, we first concatenate all the five relation-refined contexts at the token level:
\begin{align}
    \bm{H}_{Cog}[i] = \mathop{\textstyle \bigoplus}_{r \in \{\textit{xWant}, \textit{xNeed},  \textit{xIntent},  \textit{xEffect}\} } &\bm{H}_{Cog, r}[i] \label{eqn:12}\\
    \bm{H}_{Refine}[i] = \bm{H}_{Aff}[i] \ \oplus & \ \bm{H}_{Cog}[i]\label{eqn:13}
\end{align}
where $\bm{H}_{Refine}\in \mathbb{R}^{L\times 5d}$.
To highlight the more important features within the refined context representation, we apply the Sigmoid function on $\bm{H}_{Refine}$ to measure the importance of each relation-refined context for response generation. 
Then, we multiply $\bm{H}_{Refine}$ by the consequent importance scores, as done by \citet{majumder2020mime}. 
Finally, the obtained representation is passed through a Multi-Layer Perceptron (MLP) with ReLU activation, which learns how to mix the commonsense knowledge of different relations into a combined contextualized representation:
\begin{equation}
    \label{eqn:mlp}
    \widetilde{\bm{H}}_{CTX} = \mathbf{MLP}(\sigma(\bm{H}_{Refine})\odot \bm{H}_{Refine})
\end{equation}
where $\widetilde{\bm{H}}_{CTX} \in \mathbb{R}^{L\times d}$ and $\odot$ denotes element-wise multiplication.

\subsection{Response Generation}
Lastly, the target response $Y=[y_1, \dots, y_T]$ with length $T$, which has the same meaning of $u_k$ in Section \ref{subsec:formulation}, is generated by the decoder token by token:
\begin{equation}
    P(y_t|y_{<t}, C) = \mathbf{Dec}(\bm{E}_{y_{<t}}, \widetilde{\bm{H}}_{CTX})
\end{equation}
where $\bm{E}_{y_{<t}}$ denotes the embeddings of the tokens that have been generated.
Note that the cross attention to the encoder outputs is modified to the commonsense-refined contextual representation $\widetilde{\bm{H}}_{CTX}$, which has fused the information from both the context and the commonsense inferences.

\subsection{Training Objectives}

We adopt the standard negative log-likelihood (NLL) loss on the target response $Y$:
\begin{equation}
    \mathcal{L}_{nll} = - \sum_{t=1}^T \log P(y_t|C, y_{<t})
\end{equation}

\noindent \textbf {Response Diversity} \quad
In our preliminary experiments, we noticed that the models trained on our studied dataset tend to generate similarly generic empathetic responses. As shown in Table \ref{table:phrase_count}, there are phrases that are extensively repeated within the model responses in this dataset. Hence, similar to the problem raised by \citet{li2016diversitypromoting} for generic response generation in Seq2Seq models, we believe that models trained on this dataset tend to assign a higher probability to responses that include these phrases and thus, generate safe empathetic responses (e.g. \textit{I am sorry to hear that}, \textit{That is good to hear}, and \textit{Oh no that is awful}). We consider these responses safe and generic as they do not necessarily rely on nor give much information about the user's context and can be employed in many different situations.

\begin{table}[ht]
  \centering
    \begin{tabular}{c c}
        \toprule
        \textbf{Phrases} &
        \textbf{Prop. (\%)}\\
        \midrule
        \textit{That is a} / \textit{Oh no that} / \textit{To hear that}& $\geq67$ \\
        \textit{I am so} / \textit{Sorry to hear} / \textit{Wow that is} & $\geq50$ \\
        \textit{That is really} / \textit{I am sure} / \textit{That is great} & $\geq40$ \\
        \bottomrule
    \end{tabular}
  \caption{
   Most common trigrams in the training set of \textsc{EmpatheticDialogues}. Proportion represents the number of responses that include the trigram divided by the total number of responses (e.g. more than 50\% of the responses include the trigram \textit{Sorry to hear}).
  }
  \label{table:phrase_count}
\end{table}

To tackle this issue, we implement Frequency-Aware Cross-Entropy (FACE) \cite{Jiang_2019} as an additional loss to penalize high-frequency tokens using a weighting scheme. Hence, during training and prior to receiving a new batch of samples,  we first calculate the relative frequency $RF_i$ for each vocabulary token $c_i$ in the training corpus:
\begin{equation}
    RF_i = \frac{\mathrm{freq}(c_i)}{\sum^{V}_{j=1} \mathrm{freq}(c_i)}
\end{equation}
where $V$ denotes the vocabulary size. 
Accordingly, we derive the frequency-based weight $w_i$ as follows:
\begin{equation}
    w_i = a \times RF_i + 1
\end{equation}
where $a=-(\max_{1\le j \le V}(RF_j))^{-1}$ is the frequency slope and $1$ is added as the bias so that $w_i$ falls into $[0,1]$. 
Since more frequent tokens would have a higher relative frequency, the obtained weights ensure that these tokens have lower weights. 
Lastly, we normalize $w_i$ to have a mean of $1$, as done by \citet{Jiang_2019}. 
The diversity loss would then be calculated as below:
\begin{equation}
    \mathcal{L}_{div} = - \sum_{t=1}^T \sum^V_{i=1} w_i \delta_t(c_i) \log P(c_i|y_{<t}, C)
\end{equation}
where $c_i$ is a candidate token in the vocabulary and $\delta_t(c_i)$ is the indicator function, which equals to 1 if and only if $c_i = y_t$ and 0 otherwise. 
All the parameters for our proposed model are trained and optimized based on the weighted sum of the three mentioned losses.
\begin{equation}
    \label{eqn:loss}
    \mathcal{L} = \gamma_1\mathcal{L}_{nll}+\gamma_2\mathcal{L}_{emo}+ \gamma_3\mathcal{L}_{div}
\end{equation}
where $\gamma_1$, $\gamma_2$, and $\gamma_3$ are hyper-parameters that we use to control the influence of the three losses. In our experiments, we set $\gamma_1=1$, $\gamma_2=1$, and $\gamma_3=1.5$. During our analysis, we found that setting the same coefficients for all losses did not produce sufficient penalties for the generic responses. Hence, we assigned a slightly higher value to $\gamma_3$.

\section{Experiments}

\subsection{Baselines}
We selected the following baseline models for comparison:

\begin{itemize}
    \item \textbf{Transformer }\cite{vaswani2017attention}: The original Transformer, which is trained to optimize the NLL loss ($\mathcal{L}_{nll}$).
    \item \textbf{Multi-Task Transformer (Multi-TRS)} \cite{rashkin2019empathetic}: A variation of the Transformer that has an additional unit for predicting the emotion. This model is trained to jointly optimize the NLL loss ($\mathcal{L}_{nll}$) and the cross-entropy loss for emotion classification ($\mathcal{L}_{emo}$).
    \item \textbf{MoEL} \cite{lin2019moel}: A Transformer-based model that uses a decoder for each possible user emotion, referred to as \textit{listener}, and softly combines the representations from these decoders to generate a response. Therefore, each decoder is optimized to learn how to respond to one type of emotion while a meta decoder is optimized to combine their representations and generate a response.
    \item \textbf{MIME} \cite{majumder2020mime}: Another Transformer-based model that mimics the detected user emotion to a degree. In this approach, the emotions are separated into negative and positive emotions. The model initially generates mimicking and non-mimicking representations for the response based on the emotion groups and is optimized to effectively blend these representations and generate an empathetic response.
    \item \textbf{EmpDG} \cite{li2020empdg}: A multi-resolution adversarial framework that consists of an empathetic generator and an interactive discriminator. The generator produces empathetic responses based on the detected emotion while the discriminator ensures that the generated responses are consistent with the context and are also empathetic. To provide a fair comparison with our model and the other baselines, we only implement the empathetic generator in our experiments as the discriminator requires information from the future turns within the conversation.
\end{itemize}

\subsection{Implementation Details}
We implemented all the models using PyTorch\footnote{https://pytorch.org/} and used 300-dimensional pre-trained GloVE vectors \cite{pennington2014glove} to initialize the word embeddings, which were shared between the encoders and the decoders. The hidden dimension for all corresponding components were set to 300. Adam \cite{kingma2017adam} optimizer with $\beta_1=0.9$ and $\beta_2=0.98$ was used for training. The initial learning rate was set to 0.0001 and we varied this value during training according to \citet{vaswani2017attention}. All the models were trained on one single TITAN Xp GPU using a batch size of 16 and early stopping. We used a batch size of 1 and a maximum of 30 decoding steps during testing and inference.  We used the same 8:1:1 train/valid/test split as provided by \citet{rashkin2019empathetic}.

\subsection{Automatic Evaluation}
We employed Perplexity (\textbf{PPL}) and Distinct-$n$ (\textbf{Dist-$n$}) \cite{li2016diversitypromoting} as our main automatic metrics. 
PPL represents the model's confidence in its set of candidate responses, with higher confidence resulting in a lower PPL. This can be used to evaluate the general quality of the generated responses.  
Dist-$n$ measures the proportion of unique $n$-grams in the generated responses and is commonly used to evaluate generation diversity. 
In addition, since our proposed model and the baselines models (except Transformer) all perform emotion classification as a part of their training process, we also report the prediction accuracy (\textbf{Acc}). 
As \citet{liu2017evaluate} had found that word overlap-based automatic metrics such as BLEU \cite{bleu} are not appropriate for evaluating dialogue systems, we do not report such metrics. 
\begin{table}[ht]
  \centering
    \begin{tabular}{c c c c c}
        \toprule
        \textbf{Models} &
        \textbf{PPL} &
        \textbf{Dist-1} & \textbf{Dist-2}&
        \textbf{Acc (\%)}\\
        \midrule
        Transformer & 37.62 & 0.45 & 2.02& -\\ 
        Multi-TRS & 37.75 & 0.41 & 1.67 & 33.57\\ 
        MoEL & 36.93  &0.44 & 2.10& 30.62\\ 
        MIME  & 37.09 &0.47 &1.90 & 31.36 \\
        EmpDG  & 37.29  &0.46 &2.02 & 30.41 \\ 
        \midrule
        CEM & 36.11 & \textbf{0.66} & \textbf{2.99} & \textbf{39.11}\\
        w/o  Aff & 36.49 &  0.56& 2.52 & 33.76 \\
        w/o  Cog & 36.63 & 0.56 & 2.47 & 36.42\\
        w/o  Div & \textbf{35.60}& 0.48 & 1.96 &38.82 \\
        \bottomrule
    \end{tabular}
  \caption{
    Results of automatic evaluation. The best results among all models are highlighted in \textbf{bold}.
  }
  \label{table:a_results}
\end{table}

Table \ref{table:a_results} shows the automatic evaluation results. 
CEM achieves the lowest perplexity, which suggests the overall quality of our generated responses is higher than the baselines. 
In addition, our model also considerably outperforms the baselines in terms of Dist-$n$, which highlights the importance of the diversity loss. 
In terms of emotion classification, CEM had a much higher accuracy compared to the baselines, which suggests the addition of commonsense knowledge is also beneficial for detecting the user's emotion.

\subsection{Human Evaluation}
In previous work, human evaluation was conducted via two tasks: 
first, crowdsourcing workers were asked to assign a score from 1 to 5 to the generated responses based on the aspects of fluency, relevancy, and empathy; 
second, they were required to choose the better response between two models within the same context. However, the criteria for giving a score from 1 to 5 is highly likely to vary between different individuals, which results in low inter-annotator agreement and is not a suitable indicator of a model's performance. 
In addition, asking workers to choose the better responses without any guidelines and solely relying on their own preference is not satisfactory. 
This is due to the fact that each person may consider aspects that are different from what is being investigated when making their choices, which is also not a reliable indicator of user preference.

\begin{table}[ht]
  \centering
    \begin{tabular}{c c c c c c}
        \toprule
        \textbf{Comparisons} &
        \textbf{Aspects} &
        \textbf{Win} &
        \textbf{Lose} &
        $\bm{\kappa}$ \\
        \midrule
        &  Coh. &\textbf{53.6$^\ddag$}& 37.6& 0.57\\
        CEM vs. MoEL& Emp. &\textbf{52.0$^\ddag$}& 38.0& 0.57\\
        & Inf.&\textbf{61.0$^\ddag$}& 30.6& 0.51\\
        \midrule
        &  Coh. &\textbf{52.0$^\ddag$} & 42.3& 0.44\\
        CEM vs. MIME& Emp. &\textbf{50.3$^\ddag$}&41.6&0.57\\
        & Inf.&\textbf{48.6$^\text{ }\text{ }$}&45.0&0.51\\
        \midrule
         &  Coh. &\textbf{46.3$^\dag$} & 42.6& 0.52\\
        CEM vs. EmpDG& Emp. &\textbf{54.3$^\ddag$}&33.3&0.51\\
        & Inf.&\textbf{47.6$^\dag$}&43.3&0.41\\
        \bottomrule
    \end{tabular}
  \caption{
    Human evaluation results (\%). Ties are not shown. $\kappa$ denotes the inter-annotator agreement measured by Fleiss's kappa, where $0.4<\kappa<0.6$ indicates moderate agreement. ${\dag}$,${\ddag}$ represent significant improvement with $p$-value $< 0.1/0.05$ respectively (sign test).
  }
  \label{table:h_results}
\end{table}

To address these issues, we conducted an aspect-based pairwise preference test. 
That is, for a given context, we paired our model's response with a response from the baselines and asked annotators to choose the better response based on the context and the following three aspects: 
1) \textit{Coherence} (\textbf{Coh.}): which response is more coherent in content and relevant to the context; 
2) \textit{Empathy} (\textbf{Emp.}): which response shows more understanding of the user's situation and presents a more appropriate emotion; 
3) \textit{Informativeness} (\textbf{Inf.}): which response conveys more information about the context.
Then, we randomly sampled 100 response pairs and assigned three crowdsourcing workers to annotate each pair.
Ties were allowed but the annotators were encouraged to choose one of the responses. 

As shown in Table \ref{table:h_results}, CEM outperforms the baselines in all of the three aspects. 
Particularly, with the enhancement of commonsense knowledge, our model was able to produce responses that conveyed more specific and informative content and thus were more empathetic. 
We also note CEM did not significantly outperform MIME in \textit{informativeness}. 
Upon further investigation, we realized that on average, MIME tends to generate longer responses (12.8 words / response) compared to CEM (9.6 words / response). It is possibly due to some annotators considering these responses as more informative since they included more words. However, as shown by the results of the automatic evaluation (Table \ref{table:a_results}), we can observe that MIME has the second-lowest Dist-2 score, which suggests that its generated responses may follow similar patterns and have less diversity.

\begin{table*}[t!]
  \centering
    \begin{tabularx}{\textwidth}{c| l}
        \toprule
        \textbf{Emotion} & Excited\\
        \textbf{Context}& I decided a few months I was ready for a puppy, so I put the word out.\\ 
        \midrule
        \textbf{MoEL}& That is good, I am glad you did not get a dog.  \\
        \textbf{MIME}& That is so sweet! I bet you are so proud! \\
        \textbf{EmpDG}& That is awesome! I bet you are very proud of you! \\
        \midrule
        \textbf{ \textit{Affect}} & Happy, \textbf{Excited}, Hopeful\\
        \textbf{ \textit{Cognition}} & To have a pet, \textbf{To get a dog}, To have a companion\\
        \textbf{CEM}& That is \textbf{great}! Did you \textbf{get a good dog}?  \\
        \midrule
        \textbf{Gold}& Put what word? You mean published the info?\\
        \bottomrule
    \end{tabularx}
\vspace{0.2cm}
    
    \begin{tabular}{c| l}
        \toprule
        \textbf{Emotion} & Sentimental\\
        \textbf{Context}& Every time I look through older photos of my children, I am taken right back to the time they were that size, \textbf{ } \\ 
        & and what we were doing ! It is a bittersweet feeling, for sure.\\
        \midrule
        \textbf{MoEL}& I am sorry to hear that, I am sorry to hear that.\\
        \textbf{MIME}&That is so sweet. I am sure you miss them.\\
        \textbf{EmpDG}& I am so happy for you. \\
        \midrule
        \textbf{ \textit{Affect}} &\textbf{Nostalgic}, Happy, Sad\\
        \textbf{ \textit{Cognition}} &\textbf{To reminisce memories}, Look at old photos, To remember the past\\
        \textbf{CEM}&Yeah, it is always \textbf{nice to have those memories.}\\
        \midrule
        \textbf{Gold}&Oh, it sure is, I feel the same way. They grow up so fast. \\
        \bottomrule
    \end{tabular}
    \vspace{0.2cm}
    
    \begin{tabular}{c| l}
        \toprule
        \textbf{Emotion} & Guilty\\
        \textbf{Context}& \textit{Speaker:} Yeah about 10 years ago I had a horrifying experience. It was 100\% their fault but they hit the water \\
        &  barrels and survived. They had no injuries but they almost ran me off the road.\\ 
         & \textit{Listener:} Did you suffer any injuries?\\ 
         & \textit{Speaker:} No I was not hit. It turned out they were drunk. I felt guilty but realized it was his fault.\\ 
        \midrule
        \textbf{MoEL}& That is so scary! I hope you are okay! \\
        \textbf{MIME}& I am sorry to hear that. I hope you can find someone else to do that. \\
        \textbf{EmpDG}& Oh no! I am sorry to hear that. I hope you were able to get it fixed.   \\
        \midrule
        \textbf{ \textit{Affect}} & Remorseful, \textbf{Guilty}, Sorry\\
        \textbf{ \textit{Cognition}} &To be safe, \textbf{No harm done}, \textbf{No one hit}\\
        \textbf{CEM}& I am glad you \textbf{did not get hurt}. \\
        \midrule
        \textbf{Gold}& Why did you feel guilty? People really shouldn't drive drunk.\\
        \bottomrule
    \end{tabular}
  \caption{
    Case study of the generated responses by CEM and the baselines.
    }
  \label{table:case_study}
\end{table*}

\subsection{Ablation Studies}
We conducted ablation studies to verify the effectiveness of each of the components in our model. Specifically, we designed three variants of CEM: \textbf{1) w/o Aff:} the affective and affection-refined encoders are removed (Equations \ref{eqn:aff} \& \ref{eqn:7}), the affect representation is neglected in the commonsense-refined representation (Equation \ref{eqn:13}), and the hidden representation of the \texttt{[CLS]} token from the encoded context (Equation \ref{eqn:init}) is used for emotion classification; \textbf{2) w/o Cog:} the cognitive and cognition-refined encoders are removed (Equations \ref{eqn:cog} \& \ref{eqn:8}), the cognition representation is neglected in the commonsense-refined representation (Equation \ref{eqn:13}), and the MLP is replaced with a linear layer (Equation \ref{eqn:mlp}); \textbf{3) w/o Div:} the diversity loss is removed from the training objectives (Equation \ref{eqn:loss}).

The obtained results are shown in Table \ref{table:a_results}. 
Both affective and cognitive information have a considerable impact on the emotion classification accuracy, which suggests that information about both the user's emotion and their situation are necessary for correctly identifying their feelings.
In addition, we also observed that removing the diversity loss results in considerably lower Dist-$n$ scores, which indicates the effectiveness of this loss in generating more diverse responses.

\subsection{Case Study}
Table \ref{table:case_study} shows comparisons between the generated responses of CEM and the three main baselines. 
In the first case, the baselines fail to realize the meaning behind \textit{ready for a puppy}, which implies that the user wants to buy or adopt a puppy. It can be observed that MoEL dismisses this implication while the other two baselines mistake the meaning behind the phrase for \textit{being ready for an event or exam}, which may cause the user to be \textit{proud} of themselves.
By accessing external knowledge, CEM better acknowledges the user's situation and implied feelings and generates an empathetic response that covers both aspects of empathy. 
That is, by detecting that the user might be \textit{excited} and may want to \textit{get a dog}, it responds with both affective (\textit{that is great}) and cognitive (\textit{did you get a good dog?}) statements.

Similarly, in the second case, unlike the baselines, CEM successfully detects that the user is being nostalgic, happy and sad, where the latter two emotions are likely to be implied in the word \textit{bittersweet}. 
In addition, CEM realizes that the user's intent behind looking through photos of their children was to \textit{reminisce memories}, which suggests that the user enjoys having those memories.

The final case demonstrates CEM's ability to express both affective and cognitive empathy in multi-turn dialogue. 
As shown, all the baselines dismiss the user's statement \textit{I was not hit}, which implies that they are fine and no harm was done. 
In contrast, CEM correctly recognizes that there is no harm done to the user and regardless of detecting that the user might have remorse and guilt, it chooses to focus more on the important part of this situation, which is the user's health and safety.

\section{Conclusions and Future Work}
In this paper, we proposed the Commonsense-aware Empathetic Chatting Machine (CEM) to demonstrate how leveraging commonsense knowledge could benefit the understanding of the user's situation and feelings, which leads to more informative and empathetic responses. 
Our empirical automatic and manual evaluation indicated that the effectiveness of our approach in empathetic response generation. 

In the future, our work can inspire other approaches to leverage commonsense knowledge for empathetic response generation and similarly promising tasks (e.g. providing emotional support \cite{liu2021emotional}).

\section*{Acknowledgements}
This work was supported by the National Science Foundation for Distinguished Young Scholars (with No. 62125604) and the NSFC projects (Key project with No. 61936010 and regular project with No. 61876096). This work was also supported by the Guoqiang Institute of Tsinghua University, with Grant No. 2019GQG1 and 2020GQG0005.

\bibliography{aaai22}
\end{document}